\documentclass[runningheads]{llncs}
\usepackage{comment}
\usepackage{amsmath,amssymb} 
\usepackage{color}

\usepackage{multicol}
\usepackage{multirow}
\usepackage{graphicx}
\usepackage{algorithm}
\usepackage{algorithmic}

\begin{document}
\pagestyle{headings}
\mainmatter
\def\ECCVSubNumber{1059} 

\title{Segment as Points for Efficient Online Multi-Object Tracking and Segmentation} 


\titlerunning{Segment as Points for Efficient Online MOTS}
\author{Zhenbo Xu\inst{1,2}\orcidID{0000-0002-8948-1589} \and
Wei Zhang\inst{2} \and
Xiao Tan\inst{2} \and
Wei Yang \textsuperscript{1*}\orcidID{0000-0003-0332-2649}\and
Huan Huang\inst{1} \and
Shilei Wen\inst{2} \and
Errui Ding\inst{2} \and
Liusheng Huang\inst{1}}

%

\institute{University of Science and Technology of China \and
Department of Computer Vision Technology (VIS), Baidu Inc., China\\
\textsuperscript{*} Corresponding Author. E-mail: \email{qubit@ustc.edu.cn}}

\maketitle

\begin{abstract}
Current multi-object tracking and segmentation (MOTS) methods follow the tracking-by-detection paradigm and adopt convolutions for feature extraction. However, as affected by the inherent receptive field, convolution based feature extraction inevitably mixes up the foreground features and the background features, resulting in ambiguities in the subsequent instance association. In this paper, we propose a highly effective method for learning instance embeddings based on segments by converting the compact image representation to un-ordered 2D point cloud representation. Our method generates a new tracking-by-points paradigm where discriminative instance embeddings are learned from randomly selected points rather than images. Furthermore, multiple informative data modalities are converted into point-wise representations to enrich point-wise features. 
The resulting online MOTS framework, named PointTrack, surpasses all the state-of-the-art methods including 3D tracking methods by large margins (5.4\% higher MOTSA and 18 times faster over MOTSFusion) with the near real-time speed (22 FPS). Evaluations across three datasets demonstrate both the effectiveness and efficiency of our method. Moreover, based on the observation that current MOTS datasets lack crowded scenes, we build a more challenging MOTS dataset named APOLLO MOTS with higher instance density. Both APOLLO MOTS and our codes are publicly available at \url{https://github.com/detectRecog/PointTrack}.
\vspace{-2mm}
\end{abstract}

\vspace{-2mm}
\section{Introduction}
Multi-object tracking (MOT) is a fundamental task in computer vision with broad applications such as autonomous driving and video surveillance. Recent MOT methods \cite{bhat2019learning,chu2019famnet,zhang2019robust} mainly adopt the tracking-by-detection paradigm which links detected bounding boxes across frames via data association algorithms. Since the performance of association highly depends on robust similarity measurements, which is widely noticed difficult due to the frequent occlusions among targets, challenges remain in MOT especially for crowded scenes \cite{bergmann2019tracking}. More recently, the task of multi-object tracking and segmentation (MOTS) \cite{voigtlaender2019mots} extends MOT by jointly considering instance segmentation and tracking. As instance masks precisely delineate the visible object boundaries and separate adjacency naturally, MOTS not only provides pixel-level analysis, but more importantly encourages to learn more discriminative instance features to facilitate robust similarity measurements than bounding box (bbox) based methods.

Unfortunately, how to extract instance feature embeddings from segments have rarely been tackled by current MOTS methods. TRCNN \cite{voigtlaender2019mots} extends Mask RCNN by 3D convolutions and adopts ROI Align to extract instance embeddings in bbox proposals. To focus on the segment area in feature extraction, Porzi \textit{et al.} \cite{porzi2019learning} propose mask pooling to replace ROI Align. Nevertheless, as affected by the receptive field of convolutions, the foreground features and the background features are still mixed up, which is harmful for learning discriminative feature. Therefore, though current MOTS methods adopt advanced segmentation networks to extract image features, they fail to learn discriminative instance embeddings which are essential for robust instance association, resulting in limited tracking performances.


\begin{figure}[!t]
\centering
\includegraphics[width=0.7\linewidth]{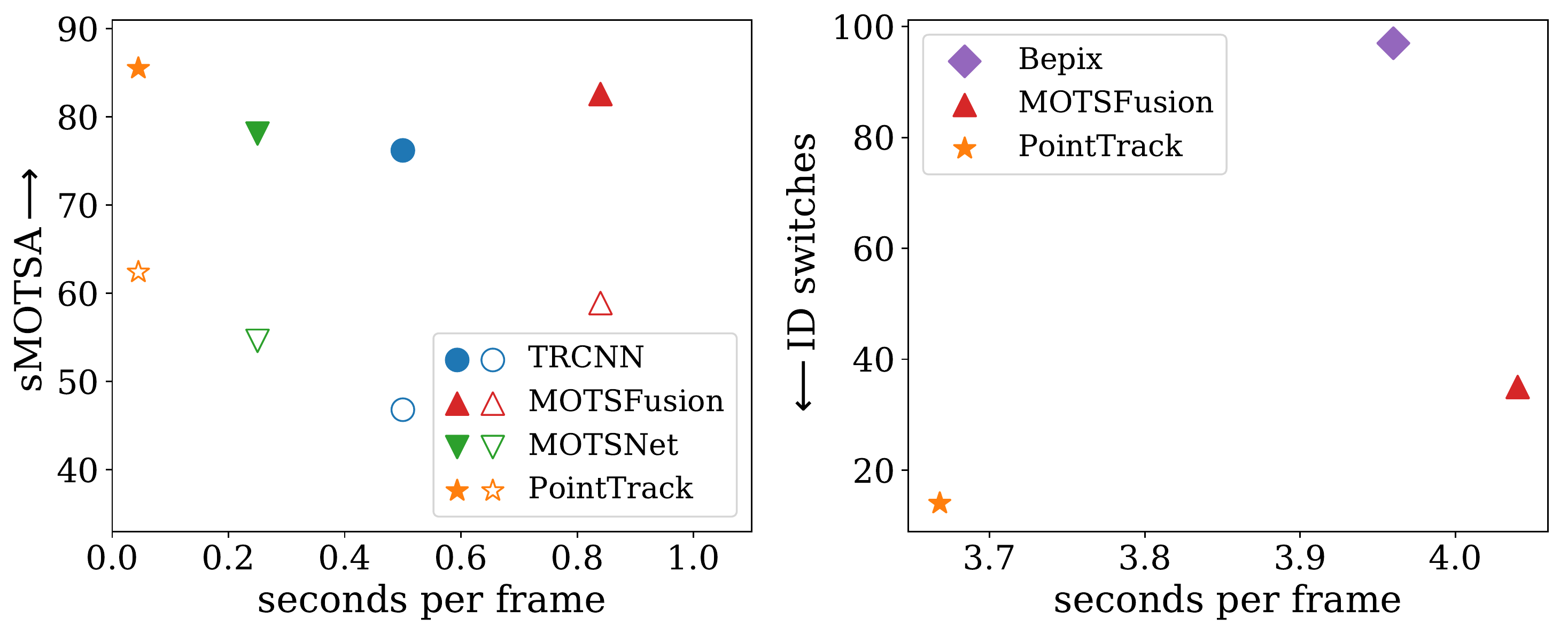}
\caption{Comparison between our PointTrack and the state-of-the-art MOTS methods on sMOTSA (Left) and id switches (Right). On the left subfigure, the filled symbols and the hollow symbols denote the results for cars and for pedestrians respectively. On the right subfigure, all methods perform tracking on the same segmentation result, which takes 3.66s.
}
\label{compare_scatter}
\vspace{-4mm}
\end{figure}


In this paper, we propose a simple yet highly effective method to learn instance embeddings on segments. Inspired by the success of PointNet \cite{qi2017pointnet} which enables feature aggregations directly from irregular formatted 3D point clouds, we regard 2D image pixels as un-ordered 2D point clouds and learn instance embeddings in a point cloud processing manner. Concretely, for each instance, we build two separate point clouds for the foreground segment and the surrounding area respectively. In each point cloud, we further propose to combine different modalities of point-wise features to realize a unified and context-aware instance embedding. In this way, the novel tracking-by-points paradigm can be easily established by combining our proposed instance embedding with any instance segmentation method.
The effectiveness of our proposed instance embedding method is examined through a comparison with current MOTS approaches based on the same segmentation results. As shown in the right subfigure of Fig. \ref{compare_scatter}, our method reduces id switches significantly. Evaluations across different datasets (see PointTrack* in Table 3,5) also prove the strong generalization ability of our proposed instance embedding.
Besides, to enable the practical utility of MOTS, we enhance the state-of-the-art one-stage instance segmentation method SpatialEmbedding \cite{Neven_2019_CVPR} for temporal coherence and build up a novel MOTS framework named PointTrack. Our proposed framework first achieves nearly real-time performance while out-performs all the state-of-the-art methods including 3D tracking methods on KITTI MOTS by large margins (see the left subfigure of Fig. \ref{compare_scatter}).



Moreover, to facilitate better evaluations, we construct a more crowded thus more challenging MOTS dataset named APOLLO MOTS based on the public ApolloScape dataset \cite{huang2018apolloscape}. APOLLO MOTS has a similar number of frames with KITTI MOTS but two times more tracks and car annotations (see Table 1). We believe APOLLO MOTS can further help promote researches in MOTS.

We summarize our main contributions as follows:
\vspace{-2mm}
\begin{itemize}
    \item We propose a highly effective method for learning discriminative instance embeddings on segments by breaking the compact image representation into un-ordered 2D point clouds.
    \item A novel online MOTS framework named PointTrack is introduced, which is more efficient and more effective than the state-of-the-art methods.
    \item We build APOLLO MOTS, a more challenging dataset with 68\% higher instance density over KITTI MOTS.
    \item Evaluations across three datasets show that PointTrack outperforms all existing MOTS methods by large margins. Also, PointTrack can reduce id switches significantly and generalizes well on instance embedding extraction.
\end{itemize}

\vspace{-2mm}
\section{Related Work}

\textbf{Tracking-by-Detection.} Detection based MOT approaches first detect objects of interests and then link objects into trajectories via data association. The data association can be accomplished on either the 2D image plane \cite{bhat2019learning,chu2019famnet,danelljan2019atom,karunasekera2019multiple,tian2019online,zhang2019robust,xu2018towards} or the 3D world space \cite{baser2019fantrack,geiger20133d,held2013precision,luo2018fast,osep2017combined,xu2020zoomnet}. ATOM \cite{danelljan2019atom} introduces a novel tracking architecture, which consists of dedicated target estimation and classification components, by predicting the overlap between the target object and an estimated bounding box. FAMNet \cite{chu2019famnet} develops an end-to-end tracking architecture where feature extraction, affinity estimation and multi-dimensional assignment are jointly optimized. Most 3D tracking methods \cite{osep2017combined,sharma2018beyond} merge track-lets based on 3D motion clues. Other approaches \cite{held2013precision,mitzel2012taking,luiten2020track} further perform 3D reconstruction for objects to improve the tracking performance.

\textbf{Tracking-by-Segmentation.} Unlike 2D bounding boxes which might overlap heavily in crowded scenes, per-pixel segments locate objects precisely. Recently instance segments have been exploited for improving the tracking performance \cite{luiten2018premvos,payer2018instance,ovsep2018track,hu2019learning,porzi2019learning}. In \cite{ovsep2018track}, Osep \textit{et al.} present a model-free multi-object tracking approach that uses a category-agnostic image segmentation method to track objects. Track-RCNN \cite{voigtlaender2019mots} extends Mask-RCNN with 3D convolutions to incorporate temporal information and extracts instance embeddings for tracking by ROI Align. MOTSNet \cite{porzi2019learning} proposes a mask pooling layer to Mask-RCNN to improve object association over time. STE \cite{hu2019learning} introduces a new spatial-temporal embedding loss to generate temporally consistent instance segmentation and regard the mean embeddings of all pixels on segments as the instance embedding for data association. As features obtained by 2D or 3D convolutions are harmful for learning discriminative instance embeddings, different from previous methods, our PointTrack regards 2D image pixels as un-ordered 2D point clouds and learn instance embeddings in a point cloud processing manner.

\textbf{MOTS Datasets.}  KITTI MOTS \cite{voigtlaender2019mots} extends the popular KITTI MOT dataset with dense instance segment annotations.
Except for KITTI MOTS, popular datasets (like the ApolloScape dataset \cite{huang2018apolloscape}) also provide video instance segmentation labels, but the instances are not consistent in time. Compared with KITTI MOTS, ApolloScape provides more crowded scenes which are more challenging for tracking. Based on this observation, we build Apollo MOTS in a semi-automatic annotation manner with the same metric as KITTI MOTS.

\vspace{-2mm}
\section{Method}

In this section, we first formulate how PointTrack converts different data modalities into a unified per-pixel style and learns context-aware instance embeddings $M$ on 2D segments. Then, details about instance segmentation are introduced. 

\vspace{-2mm}
\subsection{Context-aware instance embeddings extraction}
\begin{figure}[!t]
\centering
\includegraphics[width=1.0\linewidth]{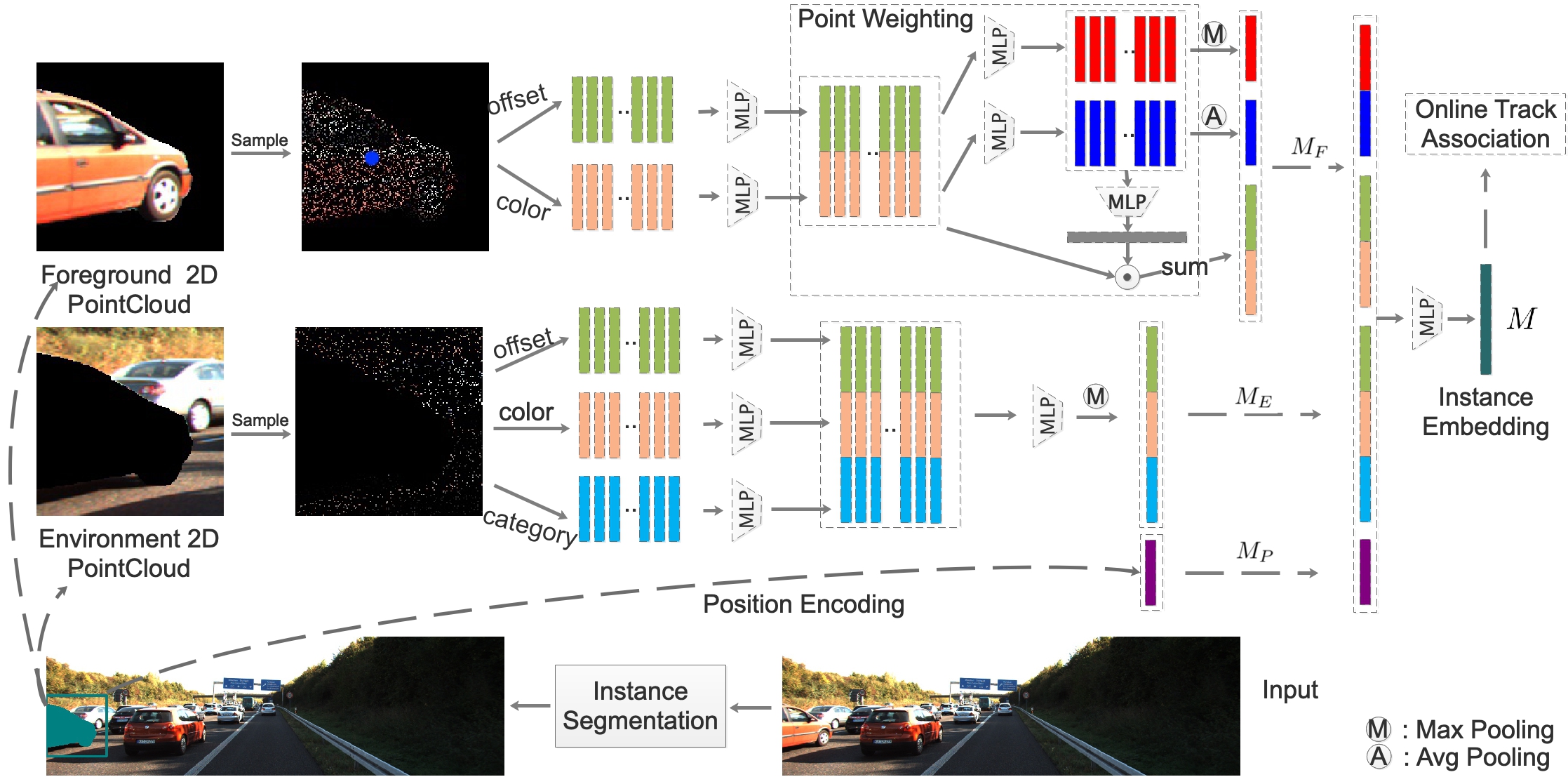}
\caption{Overview of PointTrack. For an input image, PointTrack obtains instance segments by an instance segmentation network. Then, PointTrack regards the segment and its surrounding environment as two 2D point clouds and learn features on them separately. MLP stands for multi-layer perceptron with Leaky ReLU.
}
\label{pointtrack}
\vspace{-6mm}
\end{figure}

For an instance $C$ with its segment $C_s$ and its smallest circumscribed rectangle $C_b$, we enlarge $C_b$ to $\hat{C_b}$ by extending its border in all four directions (top, down, left, and right) by a scale factor $k$ ($k=0.2$ by default). Both $C_s$ and $\hat{C_b}$ are visualized in dark green in the lower-left corner of Fig. \ref{pointtrack}. Then, we regard the foreground segment as a 2D point cloud and denote it as $F$. Similarly, we regard the other area in $\hat{C_b}$ the environment point cloud and denote it as $E$. Each point inside $\hat{C_b}$ has six dimensional data space $(u, v, R, G, B, \mathcal{C})$ that contains the coordinate $(u,v)$ in the image plane, the pixel color $(R,G,B)$, and which class $\mathcal{C}$ the pixel belongs to.

For the foreground point cloud $F$, we uniformly random sample $N_F$ points ($N_F=1000$ by default) for feature extraction. As shown in Fig. \ref{pointtrack}, $N_F$ points are enough to evenly cover a relatively large instance. For the environment point cloud $E$, $N_E$ points ($N_E=500$ by default) are randomly selected. The coordinate of the foreground point $F_i$ is denoted as $(u_i^F, v_i^F)$ and the coordinate of the environment point $E_i$ is denoted as $(u_i^E, v_i^E)$. The center point $P (u_c^F, v_c^F)$ is computed by averaging the coordinates of selected foreground points $\{F_i | i = 1, ..., N_F\}$ in the image plane. $P$ is highlighted in blue in the foreground point cloud (see Fig. \ref{pointtrack}).

Previous works \cite{sharma2018beyond,xu2019spatial,geiger20133d} have demonstrated that features concerning position, appearance, scale, shape, and nearby objects, are useful for tracking. Intuitively, PointTrack can summarize all the above features by learning the following data modalities: (i) Offset; (ii) Color; (iii) Category; (iv) Position. In the following, we formulate these data modalities and show how PointTrack learns context-aware embeddings from them. 

\textbf{Offset.} We define the offset data of each foreground point $F_i$ and each environment point $E_i$ as follows:

\begin{equation}
	O_{F_i} = (u_i^F - u_c^F, v_i^F - v_c^F), O_{E_i} = (u_i^E - u_c^F, v_i^E - v_c^F)
\end{equation}
The offset data, which are formulated as vectors from the instance center $P$ to themselves, represent the relative locations inside the segment. Offset vectors of foreground points provide essential information concerning both the scale and the shape of instances.


\textbf{Color.} We consider RGB channels and formulate the color data as follows:

\begin{equation}
	C_{F_i} = (R_i^F, G_i^F, B_i^F), C_{E_i} = (R_i^E, G_i^E, B_i^E)
\end{equation}
When the color data combine with the offset data, the discriminative appearance features can be learned from foreground points and the surrounding color distribution can be learned from environment points. The ablation study (see Table \ref{modal_impact}) shows that the color data are critical for accurate instance association.

\textbf{Category.} To further incorporate the environmental context into point-wise features, we encode all semantic class labels including the background class (suppose $Z$ classes include the background) into fixed-length one-hot vectors $\{H_j | j = 1, ..., Z\}$. Then, for selected environment points $E_i$, the one-hot category vector are also gathered for feature extraction. Suppose that $E_i$ belongs to the category $\mathcal{C}_i$, the category data are formulated as follows:

\begin{equation}
	Y_{E_i} = H_{\mathcal{C}_i}, \mathcal{C}_i \in [1, Z]
\end{equation}
Strong context features can be learned by PointTrack by jointly learning from the category data and the offset data. When the current instance is adjacent to other instances, for $E_i$ lying on the nearby instances, the category data $Y_{E_i}$ together with the offset data $O_{E_i}$ tell PointTrack both the relative position and the semantic class of nearby instances, which serve as strong clues for instance association. Visualizations (see Fig. \ref{visualization_points}) also confirm that environment points on nearby instances matter in learning discriminative instance embeddings.

\textbf{Position.} Since previous three data modalities focus on extracting features around $C_b$ regardless of the position of $C_b$ in the image plane, we encode the position of $C_b$ into the position embedding $M_P$. Following \cite{vaswani2017attention}, we embed the  position of $C_b$ ($4$-dim) into a high-dimensional vector ($64$-dim) to make it easier for learning by computing \textit{cosine} and \textit{sine} functions of different wavelengths.

Based on the above four data modalities, PointTrack learns the foreground embeddings $M_F$ and the environment embeddings $M_E$ in separate branches. As shown in Fig. \ref{pointtrack}, the environment embeddings $M_E$ are learned by first fusing $(O_{E}, C_{E}, Y_{E})$ for all $E_i$ and then applying the max pooling operation to the fused features. As aforementioned, by fusing ($O_E$, $Y_E$), PointTrack learns strong context clues concerning nearby instances from $M_E$. For the foreground point cloud $F$, $M_F$ is learned by fusing ($O_{F}$, $C_{F}$). Based on the intuition that more prominent points should have higher weights for differentiating instances, and other points should also be considered, but have lower weights, we introduce the point weighting layer to actively weight all foreground points and sum the features of all points. Different from Max-Pooling which only selects features of prominent points and Average-Pooling which blindly averages features of all points, the point weighting layer learns to summarize the foreground features by learning to weight points. Visualizations (see Fig. \ref{visualization_points}) demonstrate that the point weighting layer learns to give informative areas higher weights. Afterward, as shown in Fig. \ref{pointtrack}, $M_F$, $M_E$, and the position embeddings $M_P$ are concatenated for predicting the final instance embeddings $M$ as follows:

\begin{equation}
    M = \textbf{MLP} (M_F+M_E+M_P)
\end{equation}
where $+$ represents concatenation and \textbf{MLP} denotes multi-layer perceptron.

\textbf{Instance association.} To produce the final tracking result, we need to perform instance association based on similarities. Given segment $C_{s_i}$ and segment $C_{s_j}$, and their embeddings $M_i$ and $M_j$, the similarity $S$ is formulated as follows:
\begin{equation}
    S(C_{s_i},C_{s_j}) = -D(M_i,M_j) + \alpha * U(C_{s_i},C_{s_j})
\label{similarity}
\end{equation}
where $D$ denotes the Euclidean distance and $U$ represents the mask IOU. $\alpha$ is set to 0.5 by default. If an active track does not update for the recent $\beta$ frames, we end this track automatically. For each frame, we compute the similarity between the latest embeddings of all active tracks and embeddings of all instances in the current frame according to Eq. (\ref{similarity}). Following \cite{voigtlaender2019mots}, we set a similarity threshold $\gamma$ for instance association and instance association is allowed only when the similarity is greater than $\gamma$. The Hungarian algorithm \cite{kuhn1955hungarian} is exploited to perform instance matching. After instance association, unassigned segments will start new tracks. By default, $\beta$ and $\gamma$ are set to $30$ and $-8.0$ respectively.

\vspace{-2mm}
\subsection{Instance segmentation with Temporal Seed Consistency}

\begin{figure}[!t]
\centering
\includegraphics[width=0.8\linewidth]{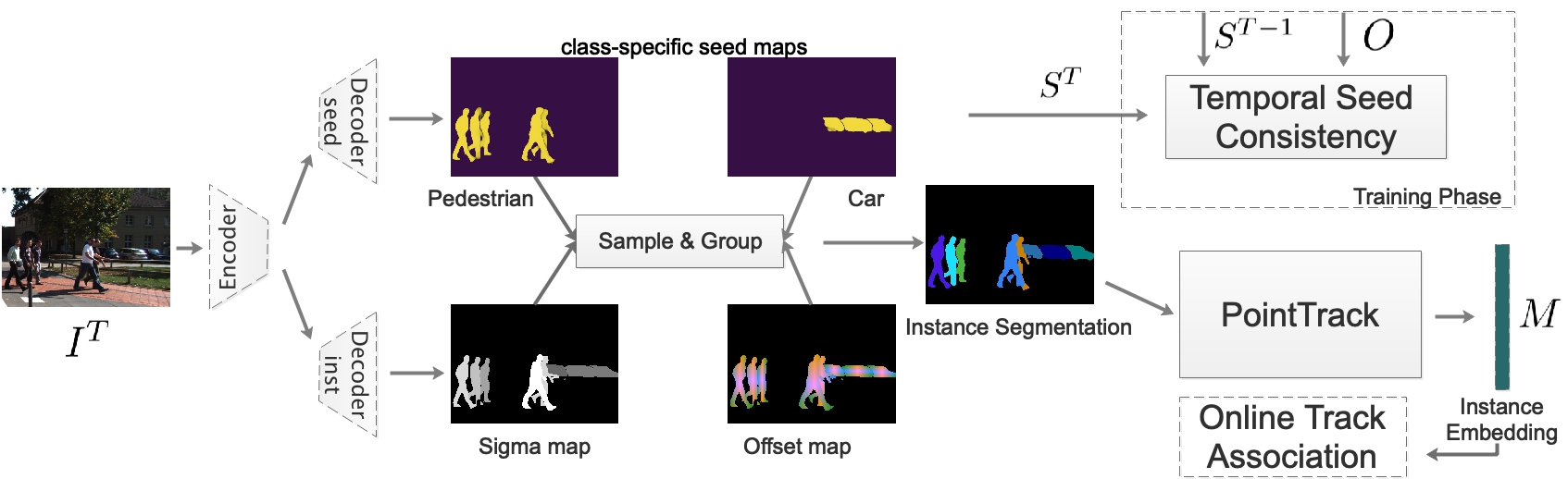}
\caption{Segmentation network of PointTrack.}
\label{SE_simple}
\vspace{-6mm}
\end{figure}

Different from previous methods \cite{voigtlaender2019mots,porzi2019learning} which put great efforts to adapt mask RCNN into the MOTS frameworks, PointTrack builds on a state-of-the-art one-stage instance segmentation method named SpatialEmbedding \cite{Neven_2019_CVPR}. SpatialEmbedding performs instance segmentation without bbox proposals, and thus runs much faster than two-stage methods. As shown in Fig. \ref{SE_simple}, SpatialEmbedding follows an encoder-decoder structure with two separate decoders: (i) the seed decoder; (ii) the inst decoder. Given an input image $I^{T}$ at time $T$, the seed decoder predicts seed maps $S^{T}$ for all semantic classes. Besides, the inst decoder predicts a sigma map denoting the pixel-wise cluster margin and an offset map representing the vector pointing to the corresponding instance center. Afterward, instance centers are sampled from $S^{T}$ and pixels are grouped into segments according to the learned clustering margin for each instance. When applied to MOTS, by studying the segmentation failure cases, we find that the seed map predictions are not consistent between consecutive frames, which results in many false positives and false negatives. Therefore, we introduce the temporal consistency loss in the training phase to improve the quality of seed map prediction as follows. First, we also feed the input image $I^{T-1}$ at time $T-1$ to SpatialEmbedding to predict the seed maps $S^{T-1}$. Then, optical flow $O$ between $I^{T-1}$ and $I^{T}$ is estimated by VCN\footnote{We exploit the pre-trained model provided at \textit{https://github.com/gengshan-y/VCN} for optical flow estimation.} \cite{yang2019volumetric}. Subsequently, we synthesize the warped seed maps $\hat{S^{T}} = O(S^{T-1})$ by exploiting $O$ to warp $S^{T-1}$. Our temporal consistency loss is formulated as:

\begin{equation}
    L_{tc} = \frac{1}{N} \sum_{i}^{N} || \hat{S_i^{T}} - S_i^{T} ||^2
\end{equation}
where $N$ is the number of foreground pixels and $i$ denotes the $i$-th foreground pixel. Evaluations (see Table 2) demonstrate that our temporal consistency loss improves the instance segmentation performance.

\vspace{-4mm}
\section{Apollo MOTS Dataset}
Tracking becomes more challenging with the increase of instances. However, for KITTI MOTS, the instance density is limited (only 3.36 cars per frame on average) and crowded scenes are also insufficient. Based on these observations, we present our Apollo MOTS dataset. We first briefly overview the dataset. Thereafter, the annotation procedures are introduced.

\subsection{Overview}
We build APOLLO MOTS on the public ApolloScape dataset \cite{huang2018apolloscape} which contains video instance segment labels for 49,287 frames. As there are barriers on both sides of the road where data were collected, pedestrians are much fewer than cars in the ApolloScape dataset. Therefore, we focus on cars. As the ApolloScape dataset calibrates the camera for each frame, APOLLO MOTS can serve as a challenging MOTS dataset for both 2D tracking and 3D tracking.\footnote{Sample videos are provided in the supplementary material.}

Detailed comparisons between our APOLLO MOTS and KITTI MOTS on their respective train/validation sets are shown in Table 1. APOLLO MOTS contains 22480 video frames including the testing set. We divide the train set, validation set, and test set according to the proportions of 30\%, 20\%, and 50\%. Scenes in these three sets have similar tracking difficulties. The original image resolution in the ApolloScape dataset is 3384 (width) x 2710 (height). We crop it to 3384 x 1604 to remove the sky area and down-sample it to a lower and more suitable resolution 1692 x 802. As shown in Table 1, though APOLLO MOTS has a similar number of frames, we have two times more tracks and car annotations. We define the car density as the number of cars per frame. The average car density of APOLLO MOTS is 5.65, which is much higher than that of KITTI MOTS. Moreover, as tracking becomes more challenging when cars are overlapped, we count the number of crowded cars for both APOLLO MOTS and KITTI MOTS. A car is considered crowded if and only if its segment is adjacent to any other car. Our APOLLO MOTS has 2.5 times more crowded cars than KITTI MOTS.

\begin{table}[!t]
\resizebox{\textwidth}{!}{%
\begin{tabular}{|l|c|c|c|c|c|c|}
\hline
 & Frames & Tracks & Annotations & Car density & Crowd cars & Frames per second \\ \hline
APOLLO MOTS & 11488 & 1530 & 64930 & 5.65 & 36403 & 10 \\ \hline
KITTI MOTS (Car) & 8008 & 582 & 26899 & 3.36 & 14509 & 7 \\ \hline
\end{tabular}
}
\label{dataset_comparison}
\caption{\textbf{Comparison between APOLLO MOTS and KITTI MOTS} on their respective train/validation sets.}
\vspace{-6mm}
\end{table}

\vspace{-4mm}
\subsection{Annotation}
We annotate all video frames in the ApolloScape dataset. If consecutive frames contain no cars or are too easy for tracking, the entire video is removed. The resulting 22480 frames represent the most difficult tracking scenes in the ApolloScape dataset. We annotate APOLLO MOTS in the following three steps.

\textbf{(1) In-complete instance segment removal.} For cars occluded by the fence, the ground-truth instance segment is always in-complete in the ApolloScape dataset. We manually traverse all frames to remove these in-complete instances by setting the bounding box area compassing this instance to the `Dontcare' category. The `Dontcare' area will be ignored in the evaluation process. After this step, only instances with complete segments will be preserved. 

\textbf{(2) Semi-automatic tracking annotation.} We incorporate PointTrack trained on KITTI MOTS into our data annotation tool for automatic instance association. For each frame, the tracking results generated by PointTrack are manually reviewed and corrected. Moreover, we subjectively assign different difficulty levels to all videos ($0\sim4$, from the easiest to the hardest) by jointly considering the crowded level, the rotation of the camera, the overlap level, etc..

\textbf{(3) Simple video removal and dataset partitioning.} All videos with difficulty level 0 are discarded. For each difficulty level from 1 to 4, we divide videos of this level to the train, validation, and test sets according to the aforementioned percentages to ensure that these sets share similar difficulties.

\section{Experiments}
Experiments are divided into four parts\footnote{More details and ablation studies are provided in the supplementary material.}. Firstly, we evaluate PointTrack across three datasets: the KITTI MOTS dataset \cite{Geiger2012CVPR}, the MOTSChallenge dataset \cite{MOT16}, and our proposed Apollo MOTS dataset. 
Secondly, we show the ablation study on data modalities. Thirdly, to investigate what PointTrack learns from 2D point clouds, we visualize both predicted instance embeddings and critical tracking points. Lastly, we provide our results on the official KITTI MOTS test set.

\textbf{Metric.} Following previous works \cite{hu2019learning,luiten2020track}, we focus on sMOTSA, MOTSA, and id switches (IDS). As an extension of MOTA, MOTSA measure segmentation as well as tracking accuracy. sMOTSA \cite{voigtlaender2019mots} is a soft version of MOTSA which weights the contribution of each true positive segment by its mask IoU with the corresponding ground truth segment.

\textbf{Experimental Setup.} Following previous works \cite{voigtlaender2019mots,hu2019learning}, we pre-train the segmentation network on the KINS dataset \cite{qi2019amodal} due to the limit of training data in KITTI MOTS (only 1704 frames contains Pedestrian where merely 1957 masks are manually annotated). Afterward, SpatialEmbedding is fine-tuned on KITTI MOTS with our proposed seed consistency loss for 50 epochs at a learning rate of $5 \cdot 10^{-6}$. For MOTSChallenge, we fine-tune the model trained on KITTI MOTS for 50 epochs at a learning rate of $5 \cdot 10^{-6}$. For APOLLO MOTS, we train SpatialEmbedding from scratch following \cite{Neven_2019_CVPR}. Besides, our PointTrack is trained from scratch in all experiments by margin based hard triplet loss \cite{yuan2019defense}. An instance database is constructed from the train set by extracting all crops $\hat{C_b}$ of all track ids. Unlike previous method \cite{voigtlaender2019mots} which samples $T$ frames as a batch, we sample $D$ track ids as a batch, each with three crops. These three crops are selected from three equally spaced frames rather than three consecutive frames to increase the intra-track-id discrepancy. The space between frames is randomly chosen between $1$ and $10$. Empirically we find a smaller $D$ ($16 \sim 24$) is better for training PointTrack, because a large $D$ (more than $40$) leads to a quick over-fitting. In addition, to test the generalization ability of instance association, we test PointTrack*, whose instance embeddings extraction is only fine-tuned on KITTI MOTS, on both MOTSChallenge and Apollo MOTS.

We compare recent works on MOTS: TRCNN \cite{voigtlaender2019mots}, MOTSNet \cite{porzi2019learning}, BePix \cite{sharma2018beyond}, and MOTSFusion (online) \cite{luiten2020track}. TRCNN and MOTSNet perform 2D tracking while BePix and MOTSFusion track on 3D. 

\begin{table}[!t]
\centering
\resizebox{1.0\textwidth}{!}{%
\begin{tabular}{|c|l|l|c|c|c|c|c|c|c|}
\hline
\multirow{2}{*}{Type} & \multicolumn{1}{c|}{\multirow{2}{*}{Method}} & \multicolumn{1}{c|}{\multirow{2}{*}{Det. \& Seg.}} & \multirow{2}{*}{Speed} & \multicolumn{3}{c|}{Cars} & \multicolumn{3}{c|}{Pedestrians} \\ \cline{5-10} 
 & \multicolumn{1}{c|}{} & \multicolumn{1}{c|}{} &  & sMOTSA & MOTSA & IDS & sMOTSA & MOTSA & IDS \\ \hline
2D & TRCNN \cite{voigtlaender2019mots} & TRCNN & 0.5 & 76.2 & 87.8 & 93 & 46.8 & 65.1 & 78 \\ \hline
3D & BePix \cite{sharma2018beyond} & RRC\cite{Ren17CVPR}+TRCNN & 3.96 & 76.9 & 89.7 & 88 & - & - & - \\ \hline
2D & MOTSNet \cite{porzi2019learning} & MOTSNet & - & 78.1 & 87.2 & - & 54.6 & 69.3 & - \\ \hline
3D & MOTSFusion \cite{luiten2020track} & TRCNN+BS & 0.84 & 82.6 & 90.2 & 51 & 58.9 & 71.9 & 36 \\ \hline
3D & BePix & RRC+BS & 3.96 & 84.9 & 93.8 & 97 & - & - & - \\ \hline
3D & MOTSFusion & RRC+BS & 4.04 & \textbf{85.5} & 94.6 & 35 & - & - & - \\ \hline
2D & PointTrack & PointTrack & \textbf{0.045} & \textbf{85.5} & \textbf{94.9} & \textbf{22} & \textbf{62.4} & \textbf{77.3} & \textbf{19} \\ \hline \hline
2D & PointTrack (without TC) & PointTrack & \textbf{0.045} & 82.9 & 92.7 & 25 & 61.4 & 76.8 & 21 \\ \hline
2D & PointTrack (on Bbox) & PointTrack & \textbf{0.045} & 85.3 & 94.8 & 36 & 61.8 & 76.8 & 36 \\ \hline
\end{tabular}%
}
\label{main_result}
\caption{\textbf{Results on the KITTI MOTS validation.} Speed is measured in seconds per frame. TC denotes the temporal consistency loss. BS represents BB2SegNet \cite{luiten2018premvos}.}
\vspace{-4mm}
\end{table}

\begin{figure}[!t]
\centering
\includegraphics[width=0.32\linewidth]{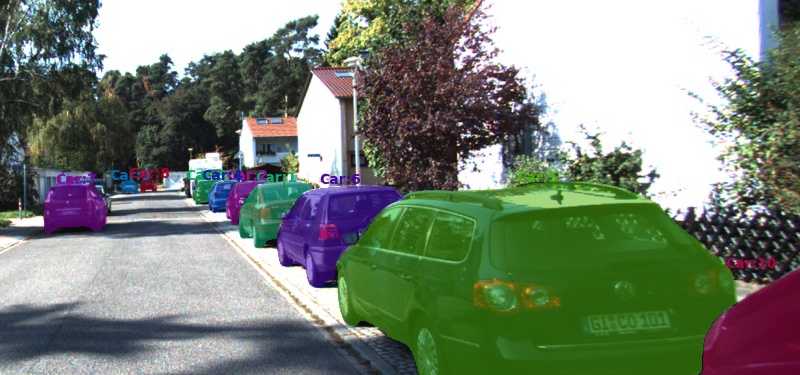}
\includegraphics[width=0.32\linewidth]{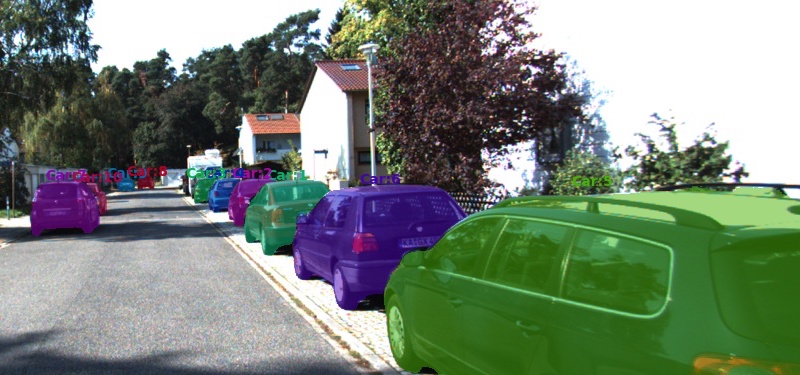}
\includegraphics[width=0.32\linewidth]{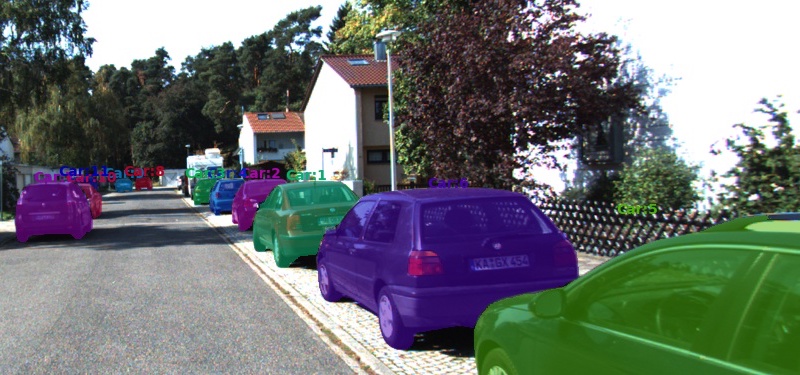}
\caption{\textbf{Quantitative results on KITTI MOTS.} Instances of the same track id are plotted in the same color.}
\label{quality}
\vspace{-6mm}
\end{figure}

\textbf{Results on KITTI MOTS.} Following MOTSFusion, we compare different methods on different segmentation results. The main results are summarized in Table 2, where our method outperforms all the state-of-the-art methods, especially for pedestrians. Quantitative results are shown in Fig. \ref{quality}. On the `Speed' column, we show the total time of detection, segmentation, and tracking\footnote{Our calculated speed is different from MOTSFusion because in \cite{luiten2020track}, the detection time of the RRC detector which takes 3.6s per frame is ignored. The speed of MOTSNet \cite{porzi2019learning} is not mentioned in their work.}. On KITTI MOTS, our PointTrack takes 0.037s per frame for instance segmentation, 8ms per frame for tracking, and 3ms per instance for embedding extraction.

For cars, the 3D tracking method MOTSFusion adopts a time-consuming detector RRC \cite{Ren17CVPR} which takes 3.6s per frame to perform detection. MOTSFusion builds up short tracklets using 2D optical flow and segments. Afterward, 3D world-space motion consistency is used to merge tracklets together into accurate long-term tracks while recovering missed detections. By contrast, though tracking objects purely on 2D images with a light-weight instance segmentation network, PointTrack achieves comparable performance to the 3D tracking method MOTSFusion (0.3\% gains on MOTSA) with significant speed improvement (0.045s VS. 4.04s). For pedestrians, PointTrack surpasses current approaches by 3.5\% and 5.4\% on sMOTSA and MOTSA respectively. It is worth noting that, though only small improvements over MOTSFusion are observed for cars on the KITTI MOTS validation set, PointTrack surpasses MOTSFusion by large margins on the official test set (see Table \ref{testset_result}), which demonstrates the good generalization ability. Besides, when the temporal consistency loss is removed (see the last but one row in Table 2), the performance drops are observable (by 2.6\% and 0.5\% on sMOTSA for cars and pedestrians, respectively). This demonstrates the effectiveness of our temporal consistency loss.


\textbf{The effectiveness of segment $C_s$.} To investigate the effectiveness of segment $C_s$, we ignore $C_s$ and instead sample points inside the inmodal bbox $C_b$. $N_E+N_F$ points are randomly sampled and the network branch for environment embedding is removed. As shown in the last row in Table 2, for cars, IDS increase by 64\% (from 22 to 36) after the segments are removed. For pedestrians, more significant performance drops (89.5\% IDS increase and 0.6\% sMOTSA) are observed. The increase in IDS demonstrates that segment matters for better tracking performances. Moreover, the gap between the performance drop in cars and in pedestrians
demonstrates that segments are more effective in improving the tracking performance for non-rigid objects where bbox level feature extraction introduces more ambiguities.


\begin{table}[!t]
\centering
\begin{tabular}{|l|c|c|c|}
\hline
\multicolumn{1}{|c|}{} & Seg. & sMOTSA & MOTSA \\ \hline
DeepSort \cite{wojke2017simple} & TRCNN & 45.71 & 57.06 \\ \hline
TRCNN \cite{voigtlaender2019mots} & TRCNN & 49.84 & 61.19 \\ \hline
DeepSort & PointTrack & 64.69 & 73.97 \\ \hline
PointTrack* & PointTrack & 70.58 & 79.87 \\ \hline
PointTrack & PointTrack & \textbf{70.76} & \textbf{80.05} \\ \hline
\end{tabular}
\caption{\textbf{Results on APOLLO MOTS validation.}}
\label{apollo_result}
\vspace{-4mm}
\end{table}

\begin{table}[!t]
\centering
\begin{tabular}{|c|c|ccc|}
\hline
Dataset & Seg. & Method & \multicolumn{1}{c|}{IDS (car)} & IDS (Ped.) \\ \hline
\multirow{5}{*}{\begin{tabular}[c]{@{}c@{}}KITTI\\ MOTS Val\end{tabular}} & \multirow{2}{*}{TRCNN} & TRCNN & 93 & 78 \\
 &  & PointTrack & \textbf{46} & \textbf{30} \\ \cline{2-5} 
 & \multirow{3}{*}{RRC+BS} & BePix & 97 & - \\
 &  & MOTSFusion & 35 & - \\
 &  & PointTrack & \textbf{14} & - \\ \hline
\multirow{5}{*}{\begin{tabular}[c]{@{}c@{}}APOLLO\\ MOTS\end{tabular}} & \multirow{3}{*}{TRCNN} & DeepSort & 1263 & - \\
 &  & TRCNN & 312 & - \\
 &  & PointTrack & \textbf{241} & - \\ \cline{2-5} 
 & \multirow{2}{*}{PointTrack} & DeepSort & 1692 & - \\
 &  & PointTrack & \textbf{292} & - \\ \hline
\multirow{2}{*}{\begin{tabular}[c]{@{}c@{}}KITTI\\ MOTS Test\end{tabular}} & \multirow{2}{*}{MOTSFusion} & MOTSFusion & 201 & 279 \\
 &  & PointTrack & \textbf{187} & \textbf{150} \\ \hline
\end{tabular}
\label{ids_result}
\caption{\textbf{Comparisons of IDS} on KITTI MOTS and APOLLO MOTS.}
\vspace{-8mm}
\end{table}

\textbf{Results on APOLLO MOTS.} We show comparisons on APOLLO MOTS in Table \ref{apollo_result}. All models are trained under the same setting as KITTI MOTS. We also train DeepSort \cite{wojke2017simple} to tracks instances on inmodal bboxes surrounding segments as a baseline to PointTrack. DeepSort extends SORT \cite{bewley2016simple} by incorporating convolution layers to extract appearance features for instance association. Different from DeepSort, our PointTrack extracts features from 2D point clouds rather than images. When applied to the same segmentation results (see the third row and the fifth row), PointTrack achieves 6\% higher sMOTSA than DeepSort and reduces IDS from 1692 to 292. Also, compared with the performance on KITTI MOTS, the sMOTSA of TRCNN and PointTrack decreases by 14.7\% and 26.4\% when evaluated on APOLLO MOTS. As the training and testing settings are the same, the significant performance drop shows that APOLLO MOTS is more challenging than KITTI MOTS.


\textbf{The significant IDS reduction by PointTrack.} As shown in Table 4, when applied to the same segmentation results on different datasets, PointTrack can effectively reduce IDS. The steady IDS reduction across different datasets and different segmentation results demonstrate the effectiveness of PointTrack. 

\begin{table}[!t]
\centering
\begin{tabular}{|l|c|c|}
\hline
 & sMOTSA & MOTSA \\ \hline
MOTDT \cite{chen2018real}+ MG & 47.8 & 61.1 \\ \hline
MHT-DAM \cite{kim2015multiple}+ MG & 48.0 & 62.7 \\ \hline
jCC \cite{keuper2018motion}+ MG & 48.3 & 63.0 \\ \hline
FWT \cite{henschel2018fusion}+ MG & 49.3 & 64.0 \\ \hline
TrackRCNN \cite{voigtlaender2019mots} & 52.7 & 66.9 \\ \hline
MOTSNet \cite{porzi2019learning} & 56.8 & 69.4 \\ \hline
PointTrack* & 57.98 & 70.47 \\ \hline
PointTrack & \textbf{58.09} & \textbf{70.58} \\ \hline
\end{tabular}%
\caption{\textbf{Results on MOTSChallenge.} +MG denotes mask
generation with a domain fine-tuned Mask R-CNN.}
\label{mcresult}
\vspace{-6mm}
\end{table}

\textbf{Results on MOTSChallenge.} Compared with KITTI MOTS, MOTSChallenge has more crowded scenarios and more different viewpoints. Following previous work \cite{voigtlaender2019mots,porzi2019learning}, we train PointTrack in a leaving-one-out fashion and show comparisons on MOTSChallenge in Table \ref{mcresult}. Our PointTrack outperforms the state-of-the-art methods by more than 1.1\% on all three metrics. It's worth noting that, though the instance embeddings extraction is only fine-tuned on KITTI MOTS (see PointTrack* in Table \ref{apollo_result},\ref{mcresult}), PointTrack* also achieves similar high performance on both APOLLO MOTS and MOTSChallenge, demonstrating the good generalization ability on instance embedding extraction.

\textbf{Ablation Study on the impact of data modalities.} We remove four data modalities in turn to examine their impacts on performance. As shown in Table \ref{modal_impact}, the largest performance drop occurs when the color data are removed. By contrast, the performance drop is minimal when the position data are removed. This difference in performance gap demonstrates that our PointTrack focuses more on the appearance features and the environment features while relies less on the bounding box position to associate instances, leading to higher tracking performances and much lower IDS than previous approaches.

\begin{table}[!t]
\centering
\begin{tabular}{|cccc|c|c|c|c|c|c|}
\hline
 &  &  &  & \multicolumn{3}{c|}{Cars} & \multicolumn{3}{c|}{Pedestrians} \\ \hline
\multicolumn{1}{|c|}{Color} & \multicolumn{1}{c|}{Offset} & \multicolumn{1}{c|}{Category} & Position & sMOTSA & MOTSA & IDS & sMOTSA & MOTSA & IDS \\ \hline
\multicolumn{1}{|c|}{$\surd$} & \multicolumn{1}{c|}{$\surd$} & \multicolumn{1}{c|}{$\surd$} & $\surd$ & \textbf{85.51} & \textbf{94.93} & \textbf{22} & \textbf{62.37} & \textbf{77.35} & \textbf{19} \\ \hline
\multicolumn{1}{|c|}{x} & \multicolumn{1}{c|}{} & \multicolumn{1}{c|}{} &  & 83.65 & 93.08 & 171 & 61.15 & 76.13 & 60 \\ \hline
\multicolumn{1}{|c|}{} & \multicolumn{1}{c|}{x} & \multicolumn{1}{c|}{} &  & 85.32 & 94.74 & 37 & 62.16 & 77.14 & 26 \\ \hline
\multicolumn{1}{|c|}{} & \multicolumn{1}{c|}{} & \multicolumn{1}{c|}{x} &  & 85.33 & 94.40 & 38 & 62.13 & 77.11 & 27 \\ \hline
\multicolumn{1}{|c|}{} & \multicolumn{1}{c|}{} & \multicolumn{1}{c|}{} & x & 85.35 & 94.77 & 35 & 62.31 & 77.29 & 21 \\ \hline
\end{tabular}
\caption{\textbf{Ablation study on the impact of different data modalities.}}
\label{modal_impact}
\vspace{-6mm}
\end{table}

\begin{figure}[!t]
\centering
\includegraphics[width=0.7\linewidth]{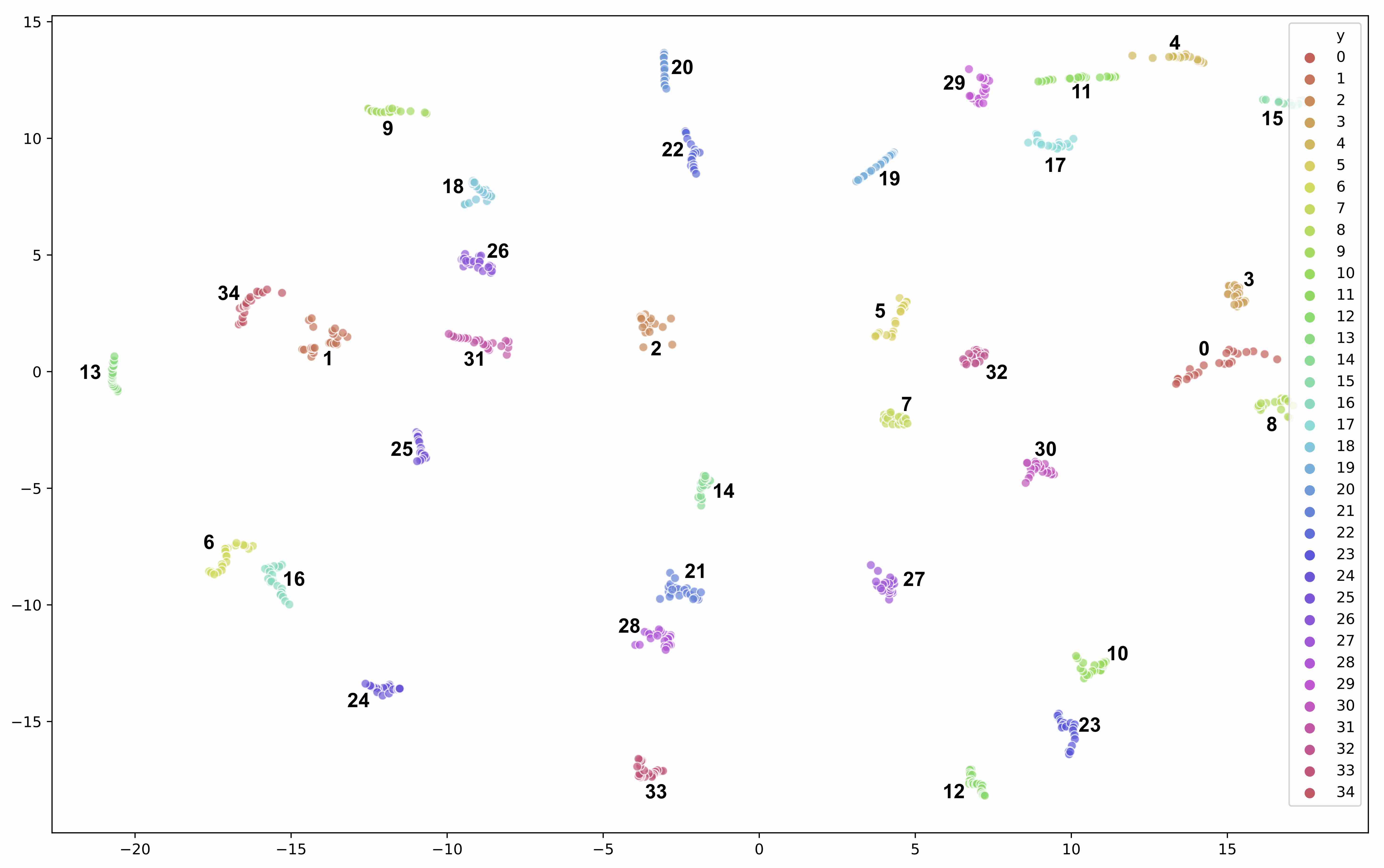}
\caption{\textbf{Visualizations of instance embeddings predicted by PointTrack.}}
\label{visualization_embedding}
\vspace{-6mm}
\end{figure}

\textbf{Visualizing instance embeddings.} We plot predicted embeddings of $20$ consecutive crops for $35$ randomly selected track ids on KITTI MOTS validation set and use t-SNE \cite{maaten2008visualizing} to embed instance embeddings ($32$-dim) into a 2D space. As shown in Fig. \ref{visualization_embedding}, embeddings belonging to different track ids are far apart, confirming that PointTrack learns discriminative instance embeddings. Surprisingly, we observe that embeddings of these track ids exhibit two distinct shapes (see Fig. \ref{visualization_embedding}): (i) Linear (index 9, 13, 20); (ii) Spherical (index 7, 26, 32). By visualization, we find that track ids in the linear shape maintain large relative movements towards the camera. They are oncoming cars or stationary cars. While for track ids in the spherical shape, most of them are cars in the same direction with small relative movements.

\begin{figure}[!t]
\centering
\includegraphics[width=0.85\linewidth]{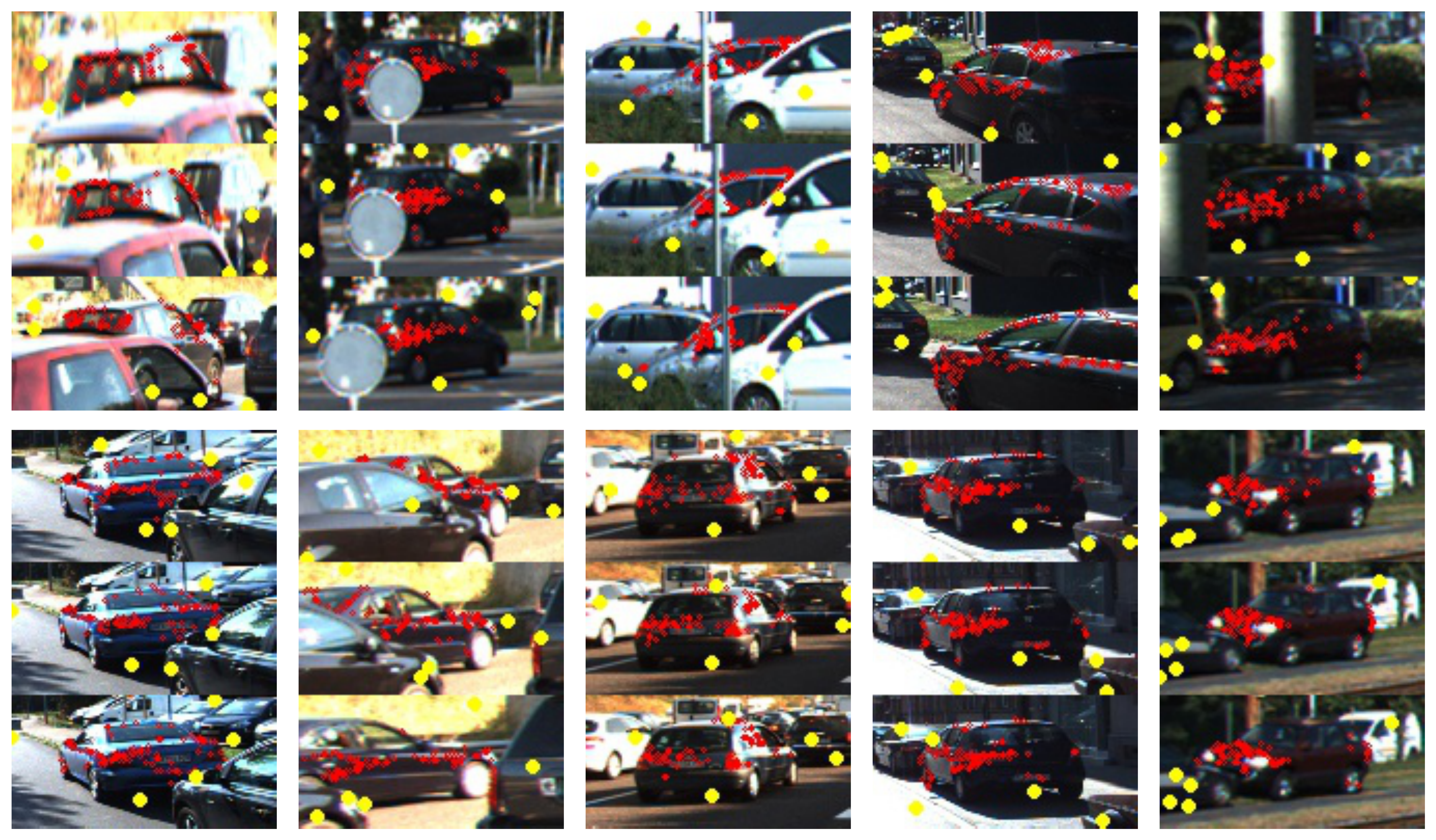}
\caption{\textbf{Visualizations of critical points.} Red points and yellow points represent the critical foreground points and the critical environment points respectively.}
\label{visualization_points}
\end{figure}

\textbf{Visualizing critical points.} We visualize critical foreground points as well as critical environment points in Fig. \ref{visualization_points}. For each instance, to validate the temporal consistency of critical points, we select crops from three consecutive frames.

For foreground points, points with $10$\% top weights predicted by the point weighting layer are plotted in red. As shown in Fig. \ref{visualization_points}, critical foreground points gather around car glasses and around car lights. We believe that the offsets of these points are essential for learning the shape and the pose of the vehicle. Also, their colors are important to outline the instance appearance and light distribution. Moreover, we find that PointTrack keeps the consistency of weighting points in consecutive frames even when different parts are occluded (the second and the fifth in the first row), or the car is moving to the image boundary (the fourth in the first row). The consistency in point weighting across frames shows the effectiveness of our point weighting layer.

For environment points, we visualize the five most critical points in yellow. These points are selected by first fetching the tensor with size of $256*N_E$ before the max-pooling layer in the environment branch and then gathering the index with the max value for all $256$-dimensions. Among these $256$ indexes, points belonging to the five most common indexes are selected. As shown in Fig. \ref{visualization_embedding}, when instances are adjacent to any other instances, yellow points usually gather on nearby instances. As aforementioned, when combining the category data with the offset data, strong context clues are provided from environment points for instance association. The distribution of critical environment points validate that PointTrack learns discriminative context features from environment points.

\begin{table}[!t]
\centering
\begin{tabular}{|l|c|c|c|c|}
\hline
 & \multicolumn{2}{c|}{Cars} & \multicolumn{2}{c|}{Pedestrians} \\ \cline{2-5} 
 & sMOTSA & MOTSA & sMOTSA & MOTSA \\ \hline
TRCNN & 67.00 & 79.60 & 47.30 & 66.10 \\ \hline
MOTSNet & 71.00 & 81.70 & 48.70 & 62.00 \\ \hline
MOTSFusion & 75.00 & 84.10 & 58.70 & 72.90 \\ \hline
PointTrack & \textbf{78.50} & \textbf{90.90} & \textbf{61.50} & \textbf{76.50} \\ \hline
\end{tabular}%
\caption{\textbf{Results on KITTI MOTS test set.}}
\label{testset_result}
\vspace{-6mm}
\end{table}

\textbf{Results on KITTI MOTS Testset.}
To further demonstrate the effectiveness of PointTrack, we report the evaluation results on the official KITTI test set in Table \ref{testset_result} where our PointTrack currently ranks first. It is worth noting that, on MOTSA, PointTrack surpasses MOTSFusion by 6.8\% for cars and 3.6\% for pedestrians. Also, PointTrack is the most efficient framework among current approaches. More detailed comparisons can be found online\footnote{The official leader-board: \textit{http://www.cvlibs.net/datasets/kitti/eval\_mots.php}}.


\vspace{-2mm}
\section{Conclusions}
In this paper, we presented a new tracking-by-points paradigm together with an efficient online MOTS framework named PointTrack, by breaking the compact image representation into 2D un-ordered point clouds for learning discriminative instance embeddings. Different informative data modalities are converted into point-level representations to enrich point cloud features. Evaluations across three datasets demonstrate that PointTrack surpasses all the state-of-the-art methods by large margins. Moreover, we built APOLLO MOTS, a more challenging MOTS dataset over KITTI MOTS with more crowded scenes.

\vspace{-2mm}
\section{Acknowledgment}
This work was supported by the Anhui Initiative in Quantum Information Technologies (No. AHY150300).

\clearpage
\bibliographystyle{splncs04}
\bibliography{egbib}
\end{document}